\documentclass[runningheads]{llncs}
\usepackage[T1]{fontenc}
\usepackage{graphicx}
\usepackage{amsmath, amssymb, mathtools, cite, csquotes, cancel}
\usepackage{aligned-overset}
\usepackage{hyperref}
\usepackage{cleveref}
\usepackage{float}
\usepackage{xcolor}
\usepackage{caption}
\usepackage{subcaption}
\newcommand{\figref}[1]{\figurename~\ref{#1}}

\allowdisplaybreaks

\usepackage[linesnumbered,ruled,vlined]{algorithm2e}
\crefname{algocf}{Algorithm}{Algorithms}

\DeclareMathOperator{\R}{\mathbb R}
\DeclareMathOperator{\N}{\mathbb N}

\DeclareMathOperator{\C}{\mathcal C}
\let\P\relax
\DeclareMathOperator{\P}{\mathcal P}

\renewcommand{\d}{\mathop{}\!\mathrm{d}}
\DeclareMathOperator{\tr}{tr}
\DeclareMathOperator{\tT}{\mathsf T}

\DeclareMathOperator{\diag}{diag}
\DeclareMathOperator{\id}{id}
\let\eps\varepsilon

\usepackage{dsfont}
\DeclareMathOperator{\1}{\mathds{1}}

\usepackage{color}

\begin{document}
\title{Accelerated Stein Variational Gradient Flow}
\titlerunning{Accelerated Stein Variational Gradient Flow}
%
\author{Viktor Stein \inst{1}\orcidID{0000-0001-9040-820X} \and
Wuchen Li\inst{2}\orcidID{0000-0002-2218-5734}}
\authorrunning{V. Stein, W. Li}

\institute{Institute of Mathematics, Technical University Berlin, 10623 Berlin, Germany,
\email{stein@math.tu-berlin.de}. 
\url{https://tu.berlin/imageanalysis}. \and
Department of Mathematics,
University of South Carolina,
SC 29225 Columbia, USA. \email{wuchen@mailbox.sc.edu}.}
\maketitle              
\begin{abstract}
    Stein variational gradient descent (SVGD) is a kernel-based particle method for sampling from a target distribution, e.g., in generative modeling and Bayesian inference.
    SVGD does not require estimating the gradient of the log-density, which is called score estimation.
    In practice, SVGD can be slow compared to score-estimation based sampling algorithms.    
    To design fast and efficient high-dimensional sampling algorithms, we introduce ASVGD, an accelerated SVGD, based on an accelerated gradient flow in a metric space of probability densities following Nesterov's method. 
    We then derive a momentum-based discrete-time sampling algorithm, which evolves a set of particles deterministically. To stabilize the particles' momentum update, we also study a Wasserstein metric regularization. 
    For the generalized bilinear kernel and the Gaussian kernel, toy numerical examples with varied target distributions demonstrate the effectiveness of ASVGD compared to SVGD and other popular sampling methods. 

\keywords{Nesterov's accelerated gradient method \and Information geometry \and Kernel methods \and Score functions \and Interacting particle system.}
\end{abstract}

\section{Introduction}
Sampling from complicated and high-dimensional target distributions $\pi \sim e^{- f}$ is essential in many applications, including Bayesian inverse problems \cite{LBADDP2022}, Bayesian neural networks \cite{N2012}, and generative models \cite{AFHHSS2021}.
Classical sampling methods are based on Markov Chain Monte Carlo (MCMC) methods \cite{H1970}.
A typical example is the unadjusted Langevin algorithm (ULA), which is a time discretization of the overdamped Langevin dynamics $\d{X}_t = - \nabla f(X_t) \d{t} + \sqrt{2} \d{B}_t$, where $(B_t)_{t \ge 0}$ is a Brownian motion.
However, ULA can have low accuracy or high variance, and requires a small step size in high dimensions and incurs a step-size dependent asymptotic bias.

Score-based models \cite{SWMG2015} take a different approach and reformulate the overdamped Langevin dynamics into the ODE $\dot{X}_t = - \nabla f(X_t) - \nabla \log\rho_t(X_t)$, where $\rho_t$ is the probability density of the $X_t$ and $\nabla \log\rho_t$ is called the score function.
The score-based ODE can be interpreted as the gradient flow \cite{AGS08} of the Kullback-Leibler (KL) divergence on the Wasserstein-2 metric space of probability measures with finite second moment \cite{JKO1998}.
Compared to ULA, the update of the score-based ODE is deterministic (or: noise-free).
A difficulty is that the score estimation is often sensitive to the chosen kernel and its bandwidth.
On the other hand, Stein variational gradient descent (SVGD) \cite{LW2016} is a score-estimation-free, kernel-based, deterministic interacting particle algorithm.
SVGD is the time discretization of the gradient flow of the KL divergence with respect to a kernelized metric in probability space, namely the Stein metric.
Since its update equation includes not only a drift term, but also an interaction term, it often only few particles suffice to attain good exploration of the target distribution.
In practice, SVGD can converge quite slowly. 

Inspired by Nesterov's accelerated gradient descent method \cite{N1983}, the authors of \cite{TM2019,WL2022,CLTW2025} introduced accelerated information gradient (AIG) flows, which generalize Nesterov's acceleration dynamics in metric spaces of probability measures.
A natural question arises: {\em What is the accelerated SVGD and how can we update the momentum variable?}

In this paper we introduce ASVGD, an accelerated sampling algorithm based on SVGD and AIG flow.
We study a kernelized accelerated gradient flow in the probability space equipped with the Stein metric regularized by the Wasserstein-2 metric.
We approximate the AIG flow using the interacting particles' positions and momenta.
This algorithm avoids score estimation by using integration by parts to shift the derivative of the log-density onto the kernel function.

This paper is organized as follows. In section \ref{sec:metricGradientFlows}, we review metric gradient flows in the density manifolds equipped with either the Wasserstein or the Stein metric.
Next, we derive the accelerated Stein gradient flow and its associated particle algorithm in section \ref{sec3}.
Several numerical examples with the generalized bilinear kernel and the Gaussian kernel are presented in section \ref{sec:numerics}.

\section{Metric gradient flows on the density manifold} \label{sec:metricGradientFlows}
In this section, we briefly review gradient flows on the density manifold. 
Let $\Omega \subset \R^d$.
We assume that either $\Omega$ has periodic boundary conditions, or that $\Omega = \R^d$.
Instead of considering all probability measures on $\Omega$, we work exclusively on its subset of smooth positive probability densities,
\begin{equation*}
    \widetilde{\P}(\Omega) \coloneqq \left\{ \rho \in \C^{\infty}(\Omega): \rho(x) > 0, \forall x \in \Omega, \int_{\Omega} \rho(x) \d{x} = 1 \right\}.
\end{equation*}
The tangent space to $\widetilde{\P}(\Omega)$ is $T_{\rho} \widetilde{\P}(\Omega) \coloneqq \left\{ \sigma \in \C^{\infty}(\Omega): \int_{\Omega} \sigma(x) \d{x} = 0 \right\}$, and the cotangent space at $\rho$ is $T_{\rho}^* \widetilde{\P}(\Omega) \coloneqq \C^{\infty}(\Omega) / \R$.

\begin{definition}[Metric tensor field on $\widetilde{\P}(\Omega)$]
    A \textit{metric tensor field on $\widetilde{\P}(\Omega)$} is a smooth map $G \colon \rho \mapsto G_{\rho}$ on $\widetilde{\P}(\Omega)$ such that $G_{\rho} \colon T_{\rho} \widetilde{\P}(\Omega) \to T_{\rho}^* \widetilde{\P}(\Omega)$ is smooth and invertible for each $\rho \in \widetilde{\P}(\Omega)$.    
\end{definition}

A metric tensor field yields a \textit{metric $g$ on $\widetilde{\P}(\Omega)$} via
\begin{equation*}
     g_{\rho}
     \colon T_{\rho} \widetilde{\P}(\Omega) \times T_{\rho} \widetilde{\P}(\Omega) \to \R, \quad
     (\sigma_1, \sigma_2) \mapsto \int_{\Omega} \sigma_1(x) \left(G_{\rho}[\sigma_2]\right)(x) \d{x}.
\end{equation*}

In $\widetilde{\P}(\Omega)$, the differential of a functional $E \colon \widetilde{\P}(\Omega) \to \R$ can conveniently be expressed using the metric tensor field and the functional derivative. 

\begin{definition}[First linear functional derivative]
    If it exists, its first functional derivative  of $E$ is the one-form $\delta E \colon \widetilde{\P}(\Omega) \to \C^{\infty}(\Omega) / \R$ with 
    \begin{gather*}
        \langle \delta E(\rho), \phi \rangle_{L^2(\Omega)}
        = \frac{\d}{\d{t}}\bigg|_{t = 0} E(\rho + t \phi), \quad \forall \phi \in \C^{\infty}(\Omega): \rho + t \phi \in \widetilde{\P}(\Omega), | t | \text{ small}.
    \end{gather*}
\end{definition}

\begin{definition}[Metric gradient flow on $\widetilde{\P}(\Omega)$]
    A smooth curve $\rho \colon [0, \infty) \to \widetilde{\P}(\Omega)$, $t \mapsto \rho_t$ is a $(\widetilde{\P}(\Omega), G)$-gradient flow of $E$ starting at $\rho(0)$ if 
    \begin{equation} \label{eq:metric_GF}
        \partial_t \rho_t
        = - G_{\rho_t}^{-1}[\delta E(\rho_t)], \qquad \forall t > 0.
    \end{equation}
\end{definition}
From now on we will consider only the energy given by the KL divergence $E(\rho) = \mathrm{D}_{\mathrm{KL}}(\rho \| \pi) = \int_{\Omega} \rho(x) \log\frac{\rho(x)}{\pi(x)} \d{x}$ with the target distribution $\pi = Z^{-1} e^{- f} \in \widetilde{\P}(\Omega)$, which has the smooth \textit{potential} $f \colon \Omega \to \R$ and finite normalization constant $Z \coloneqq \int_{\Omega} e^{-f} \d{x}$.
We have $\delta E(\rho) = \log\frac{\rho}{\pi} + 1$. Hence
\begin{equation*}
    \nabla \delta E(\rho)
    = \nabla \log\rho - \nabla \log\pi
    = \nabla \log\rho + \nabla f.  
\end{equation*}

We focus on two examples of density manifolds, using the Wasserstein-2 metric \cite{V2003} and the Stein metric \cite{LW2016,L2017,NR2023}.
\begin{example}[Wasserstein-2 metric gradient flow]
    The Wasserstein metric is defined via the inverse metric tensor field
    \begin{equation*}
        [G_\rho^{W}]^{-1} \colon T_{\rho}^* \widetilde{\P}(\Omega) \to T_{\rho} \widetilde{\P}(\Omega), \qquad
        \Phi \mapsto - \nabla \cdot (\rho \nabla \Phi), \qquad \rho \in \widetilde{\P}(\Omega).
    \end{equation*}
   The $(\widetilde{P}(\Omega), G^{W})$-gradient flow of $E$ is $$\partial_t \rho_t
   = \nabla \cdot( \rho_t (\nabla \log\rho_t + \nabla f))
   = \nabla\cdot(\nabla \rho_t )+ \nabla \cdot (\rho_t \nabla f),$$ where we used the fact that $\rho_t \nabla \log \rho_t = \nabla \rho_t.$
\end{example}

\begin{example}[Stein metric gradient flow] \label{example:SteinMetric}
    For a symmetric, positive definite \cite{W2004} and smooth kernel $K \colon \Omega \times \Omega \to \R$, the Stein metric is defined via the inverse metric tensor field
    \begin{equation*}
        (G_{\rho}^{(K)})^{-1}(\Phi) \coloneqq \left( x \mapsto - \nabla_x \cdot \left( \rho(x) \int_{\Omega} K(x, y) \rho(y) \nabla \Phi(y) \d{y} \right) \right).
    \end{equation*}
    The $(\widetilde{\P}(\Omega), G^{K})$-gradient flow of $E$ is
    \begin{align*}
        \partial_t \rho_t(x)
        & = \nabla_x \cdot \left( \rho_t(x) \int_{\Omega} K(x, y) \rho_t(y) (\nabla \log\rho_t(y) + \nabla f(y)) \d{y} \right) \\
        & = \nabla_x \cdot \left(\rho_t(x) \int_{\Omega} \left( K(x, y) \nabla f(y) - \nabla_2 K(x, y)\right) \rho_t(y) \d{y}\right).
    \end{align*}
   Here we used $\rho_t(y)\nabla\log\rho_t(y)=\nabla\rho_t(y)$ and applied the integration by parts for the score function $\nabla\log\rho_t(y)$.   
\end{example}

\section{Accelerated Stein variational gradient flows on the density manifold}\label{sec3}

Generalizing the continuous limit of Nesterov's accelerated gradient descent \cite{N1983} to the density manifold $\widetilde{\P}(\Omega)$ yields the following definition.

\begin{definition}[Accelerated $(\widetilde{\P}(\Omega), G)$-gradient flow]
    Let $\alpha \colon \R_{\ge 0} \to \R_{\ge 0}$ be a \enquote{damping}.
    The $\alpha$-accelerated $(\widetilde{\P}(\Omega), G)$-gradient flow of $E \colon \widetilde{\P}(\Omega) \to \R$ is the curve $(\rho_t)_{t > 0}$ solving the Hamiltonian flow \cite[Chp.~X]{B2022} with an added linear damping term, that is,
    \begin{equation} \label{eq:acc_Hamiltonian_flow}
        \partial_t \begin{pmatrix}
            \rho_t \\ \Phi_t
        \end{pmatrix}
        + \begin{pmatrix}
            0 \\ \alpha_t \Phi_t
        \end{pmatrix}
        - \begin{pmatrix}
            0 & 1 \\ -1 & 0
        \end{pmatrix} 
        \begin{pmatrix}
            \delta_1 H(\rho_t, \Phi_t) \\
            \delta_2 H(\rho_t, \Phi_t)
        \end{pmatrix}
        = 0,      
    \end{equation}
    with $\rho(0) = \rho_0$ and $\Phi(0) = 0$, where 
    \begin{equation*}
        H \colon T^* \widetilde{\P}(\Omega) \to \R, \qquad 
        (\rho, \Phi) \mapsto \frac{1}{2} g_{\rho}(G_{\rho}^{-1}[\Phi], G_{\rho}^{-1}[\Phi]) + E(\rho),
    \end{equation*}
    is the Hamiltonian, and $\delta_i$ for $i \in \{ 1, 2 \}$ denotes the functional derivative with respect to the $i$-th component.
\end{definition}
By \cite[Prop.~1]{WL2022}, \eqref{eq:acc_Hamiltonian_flow}
for $G = G^{(K)}$ becomes 
\begin{equation} \label{eq:Accelerated_Stein_Flow}
    \begin{cases}
        \partial_t \rho_t + \nabla \cdot \left(\rho_t \int_{\Omega} K(\cdot, y) \rho_t(y) \nabla \Phi_t(y) \d{y}\right)
        = 0, \\
        \partial_t \Phi_t + \alpha_t \Phi_t 
        + \int_{\Omega} K(y, \cdot) \langle \nabla \Phi_t(y), \nabla \Phi_t(\cdot) \rangle \rho_t(y) \d{y} + \delta E(\rho_t)
        = 0.
    \end{cases}
\end{equation}
We consider the initial distribution $\rho_0$ with initial potential function $ \Phi_0 = 0$.

\subsection{Particle algorithms}

By formally replacing $\rho_{t}$ in \eqref{eq:Accelerated_Stein_Flow} by its empirical estimation $\frac{1}{N} \sum_{j = 1}^{N} \delta_{X_{t}^{j}}$ using the Dirac measures $\delta$, we can simulate \eqref{eq:Accelerated_Stein_Flow} using $N$ particles $(X_{t}^{j})_{j = 1}^{N} \subset \R^d$ and their accelerations $(Y_t^{j})_{j = 1}^{N} \subset \R^d$ at time $t$.
We then employ a forward Euler discretization in time.

In \cite{WL2022}, the following deterministic particle discretization was introduced:
\begin{equation*}
    \begin{cases}
        X_{j}^{k+1}
        = X_{j}^{k} + \frac{\sqrt{\tau}}{N} \sum_{i = 1}^{N} K(X_{j}^{k}, X_{i}^{k}) V_{i}^{k}, \\
        V_{j}^{k+1}
        = \alpha_{k} V_{j}^{k}
        - \frac{\sqrt{\tau}}{N} \sum_{i = 1}^{N} (\nabla_1 K)(X_{j}^{k}, X_{i}^{k}) \langle V_{j}^{k}, V_{i}^{k} \rangle
        - \sqrt{\tau} \left( \nabla f(X_{j}^{k}) + \xi_{j}^{k}\right),
    \end{cases}
\end{equation*}
for $j \in \{ 1, \ldots, N \}$, where $\xi_{j}^{k}$ is the Gaussian KDE of the score evaluated $X_{j}^{k}$ and $\tau > 0$ is the step size.
The KDE is very sensitive to the kernel width, which is selected using the Brownian motion method.

In this paper, we use the particle momentum $Y \colon (0, \infty) \to \R^d$, $t \mapsto \dot{X}_t$,
\begin{equation} \label{eq:Acc_Stein}
    \dot{X}_t=Y_t  = \int_{\Omega} K(X_t, y) \nabla \Phi_t(y) \rho_t(y) \d{y}.
\end{equation}
\begin{lemma}[Accelerated Stein variational gradient flows with particles' momenta]
    The deterministic interacting particle system associated to \eqref{eq:Accelerated_Stein_Flow} is
    \begin{equation} \label{eq:dot_Yt}
        \begin{cases}
            \dot{X}_t = Y_t, \\
            \dot{Y}_t = - \alpha_t Y_t + \int_{\Omega} \left( K(X_t, y) \nabla f(y) - \nabla_2 K(X_t, y)\right) \rho_t(y) \d{y} \\
            \qquad + \int_{\Omega^2} \rho_t(y) \rho_t(z) \langle \nabla \Phi_t(z), \nabla \Phi_t(y) \rangle\cdot \bigg[K(y, z) (\nabla_2 K)(X_t, y)  \\
            \qquad \qquad + K(X_t, z) (\nabla_1 K)(X_t, y) - K(X_t, y) (\nabla_2 K)(z, y) \bigg] \d{y} \d{z}.
        \end{cases}
    \end{equation}
\end{lemma}

Due to the page limit, we only sketch the proof.
\begin{proof}
    Plugging \eqref{eq:Accelerated_Stein_Flow} into \eqref{eq:Acc_Stein} yields for $t > 0$
    \begin{align*}
        \dot{Y}_t
        &{=} \int_{\Omega} \int_{\Omega} (\nabla_1 K)(X_t, y) K(X_t, z)  \left\langle \nabla \Phi_t(z), \nabla \Phi_t(y) \right\rangle \rho_t(z) \rho_t(y) \d{y} \d{z} \\
        & \qquad - \alpha_t  \int_{\Omega} K(X_t, y) \nabla \Phi_t(y)  \rho_t(y) \d{y}
        - \int_{\Omega} K(X_t, y) \nabla \delta E(\rho_t)(y) \rho_t(y) \d{y} \\
        & \qquad - \int_{\Omega} \int_{\Omega} K(X_t, y) \rho_t(y) \nabla_y \left[ K(z, y) \langle \nabla \Phi_t(z), \nabla \Phi_t(y) \rangle \right] \rho_t(z) \d{y} \d{z} \\
        & \qquad + \int_{\Omega} \int_{\Omega} \nabla_y \left[ K(X_t, y) \langle \nabla \Phi_t(y), \nabla \Phi_t(z) \rangle \right] \rho_t(y) K(y, z) \rho_t(z) \d{y} \d{z}. 
    \end{align*}
    In the above formula, we use the integration by parts and cancel several terms.
    We also use the fact that $E$ is the KL divergence and \cref{example:SteinMetric}.
\end{proof}

We consider two integrally strictly positive definite kernels.
For a symmetric positive definite matrix  $A \in \R^{d \times d}$, we consider the generalized bilinear kernel, $K(x, y) \coloneqq x^{\tT} A y + 1$ and the Gaussian kernel $K(x, y) \coloneqq \exp\left(-\frac{1}{2 \sigma^2} \| x - y \|_2^2\right)$ with bandwidth $\sigma^2 > 0$.

The ASVGD algorithm for both of these kernels is summarized in \cref{algo:ASVGD}, where for conciseness of notation we will use matrices $X, Y, V \in \R^{d \times N}$, whose rows are the particles $X^{i}$, $Y^{i}, V^{i}$.

To prevent the momentum becoming too large and the particles \enquote{overshooting} the target, we use the speed restart technique from \cite[Sec~5.1]{SBC2016}, applied to each particle individually.
We also use the gradient restart from \cite[Sec.~5.2]{WL2022}, which is applied to all particles simultaneously: we set $\alpha_k \gets 0$ for all $k \in \{ 1, \ldots, N \}$ if
\begin{equation*}
    - \partial_t E(\rho_t)
    = N^{-2} \sum_{i, j = 1}^{N} \langle V^j, \nabla f(X^i) + X^i - X^j \rangle K(X^i, X^j)
    < 0.
\end{equation*}

\begin{algorithm}
    \caption{Accelerated Stein variational gradient descent} \label{algo:ASVGD}
    \KwData{Number of particles $N \in \N$, number of steps $M \in \N$, step sizes $\tau > 0$, target score function $\nabla f \colon \R^d \to \R^d$. Either a symmetric positive definite matrix $A \in \R^{n \times n}$ for bilinear kernel or a bandwidth $\sigma^2 > 0$ for Gaussian kernel, regularization parameter $\eps \ge 0$, (constant damping $\beta \in (0, 1)$).}
    \KwResult{Matrix $X^{M}$, whose rows are particles that approximate the target distribution $\pi \sim \exp(-f)$.}
    \textbf{Step 0.} Initialize $Y^0 = 0 \in \R^{N \times d}$ and  \texttt{restart\_count}$=\1_N$. \\
    \For{$k=0,\ldots, M - 1$}{
    \emph{\textbf{Step 1.} Update particle positions using particle momenta.}
    $$\displaystyle X^{k + 1}
    \gets X^{k} + \sqrt{\tau} Y^{k}.$$
    \emph{\textbf{Step 2.} Form kernel matrix and update momentum in density space.}
    $$
    K^{k + 1}
    = \big(K(X_i^{k + 1}, X_j^{k + 1}) \big)_{i, j = 1}^{N},
    \qquad
    V^{k + 1}
    \gets N (K^{k + 1} + \eps \id_N)^{-1} Y^{k}.$$\\
    \emph{\textbf{Step 3.} Update damping parameter using speed and gradient restart.} \\
    \For{$i = 1, \ldots, N$}
    {\eIf{$\| X_{i}^{k + 1} - X_{i}^{k} \|_2 < \| X_{i}^{k} - X_{i}^{k - 1} \|_2$}
    {\texttt{restart\_count}$_i= 1$}
    {\texttt{restart\_count}$_i += 1$}
    } 
    Only for the Gaussian kernel: \\
    \If{$\tr\left((V^{k + 1})^{\tT} (K^{k + 1} \nabla f(X^{k + 1}) + \big(K^{k + 1} - \diag(K^{k + 1} \1_N)\big) X^{k + 1})\right) < 0$,}{\texttt{restart\_count} $= \1_N$}
    $\alpha_{i}^k = \displaystyle\frac{\texttt{restart\_count}_i - 1}{\texttt{restart\_count}_i + 2}$, $i \in \{ 1, \ldots, N \}$. \\
    Alternatively, set a \textit{constant damping} for each particle: $\alpha_i^k = \beta$. \\
    \emph{\textbf{Step 4.} Update momenta.} \\
    For the bilinear kernel:
    \begin{align*}
        Y^{k + 1}
        & \gets \alpha^{k} Y^{k}
        - \frac{\sqrt{\tau}}{N} K^{k + 1} \nabla f(X^{k + 1}) \\ 
        & \qquad + \sqrt{\tau} \left(1 + N^{-2} \tr\left((V^{k + 1})^{\tT} K^{k + 1} V^{k}\right)\right) X^{k + 1} A.   
    \end{align*}
    For the Gaussian kernel: 
    \begin{align*}
        W^{k + 1}
        & \gets N K^{k + 1} + K^{k + 1} (V^{k + 1} (V^{k + 1})^{\tT}) \circ K^{k + 1}  \\
        & \qquad - K^{k + 1} \circ (K^{k + 1} V^{k + 1} (V^{k + 1})^{\tT}), \\
        Y^{k + 1}
        & \gets \alpha^{k} Y^{k} - \frac{\sqrt{\tau}}{N} K^{k + 1} \nabla f(X^{k + 1}) \\
        & \qquad + \frac{\sqrt{\tau}}{2 N^2 \sigma^2} \left(\diag(W^{k + 1} \1_N) - W^{k + 1}\right) X^{k + 1}.
    \end{align*}
    }
\end{algorithm}

\begin{remark}[Wasserstein metric regularization]
    The Gaussian kernel matrix is invertible if the particles are distinct.
    Since its inversion can become ill-conditioned, we instead invert $K^{k + 1} + \eps \id_N$ for some small $\eps > 0$.
    This corresponds to adding the Wasserstein metric regularization $\eps [G^{(W)}]^{-1}$ to the Stein metric and adding $\eps V_t$ to the right side of \eqref{eq:Acc_Stein}.
    We ignore the $\eps$-terms in the $Y$-update, since otherwise the update is much more difficult to implement without resorting to score estimation.
\end{remark}

\section{Numerical examples}\label{sec:numerics}

We choose $\Omega = \R^2$ and compare our algorithm, ASVGD\footnote{The python code for reproducing these experiments is available online: \url{https://github.com/ViktorAJStein/Accelerated_Stein_Variational_Gradient_Flows}.}, with ULA, SVGD, and MALA (Metropolis-adjusted Langevin algorithm) which augments ULA by a Metropolis-Hastings acceptance step to remove its asymptotic bias \cite{B1994}.
Since our method is of second order in time, we also compare it to underdamped Langevin dynamics (ULD) with unit mass and friction.

In all our plots, the initial particles are blue circles, the red squares are the final particles and the lines are the trajectories between them.
The black lines represent the level lines of the target density.
In all experiments, we choose $N = 500$ particles and perform 1000 steps with step size, bandwidth, and regularization parameter $\tau = \sigma = \eps = 0.1$.
Throughout, we observe that ASVGD converges faster than SVGD.

\textit{Bilinear kernel.}
For $A = \id$ and a zero-mean initialization and a Gaussian zero-mean target, the particles will remain Gaussian \cite[Thm.~3.5]{LGBP2024}.
Empirically, we observe that even for arbitrary $A$ and non-Gaussian targets, the overall shape of the particles is always approximately a linear transformation of the initial particles, so that the target might not be matched at all.
\begin{figure}[H]
    \includegraphics[height=2.1cm]{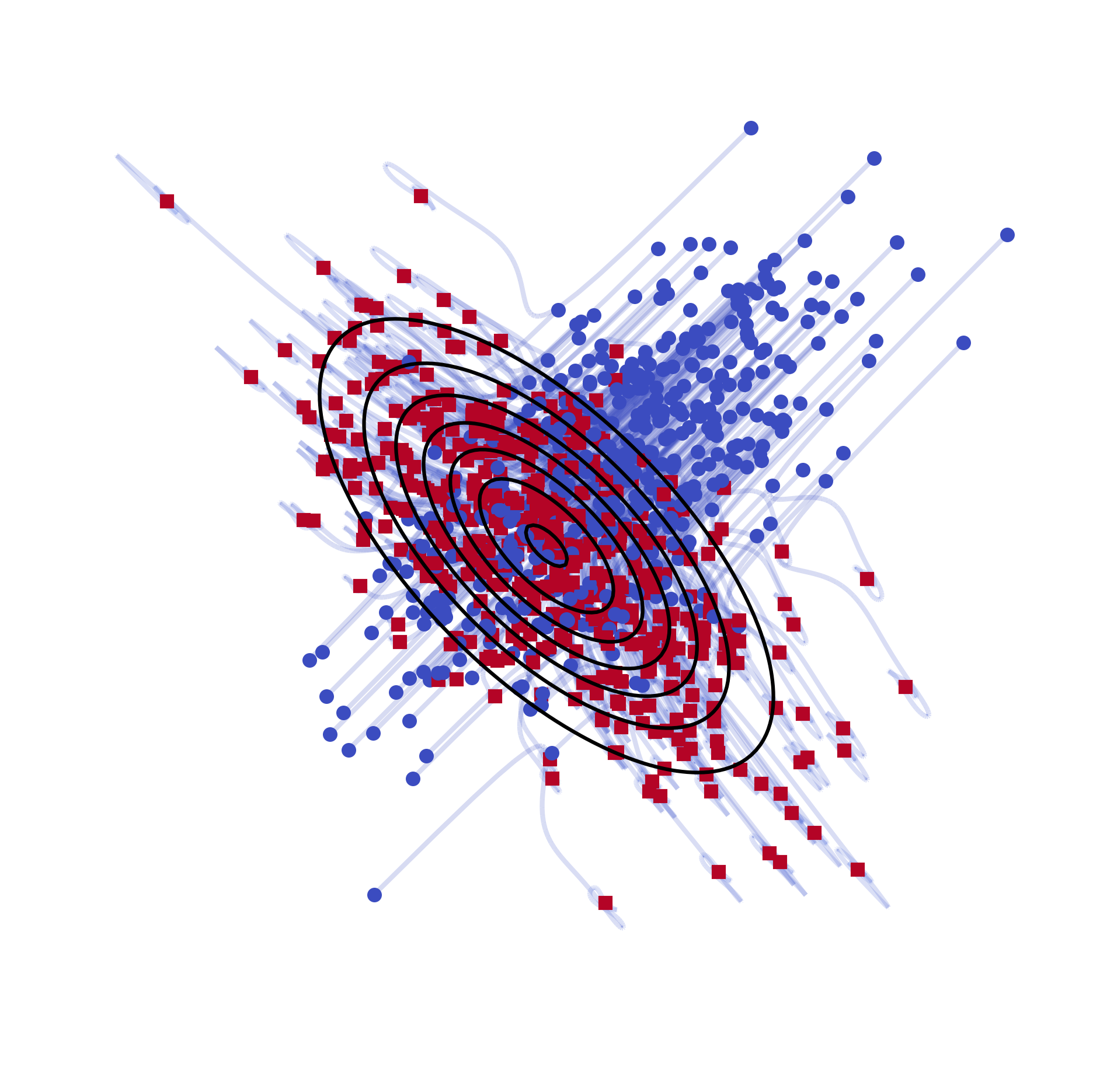}%
    \includegraphics[height=2.1cm]{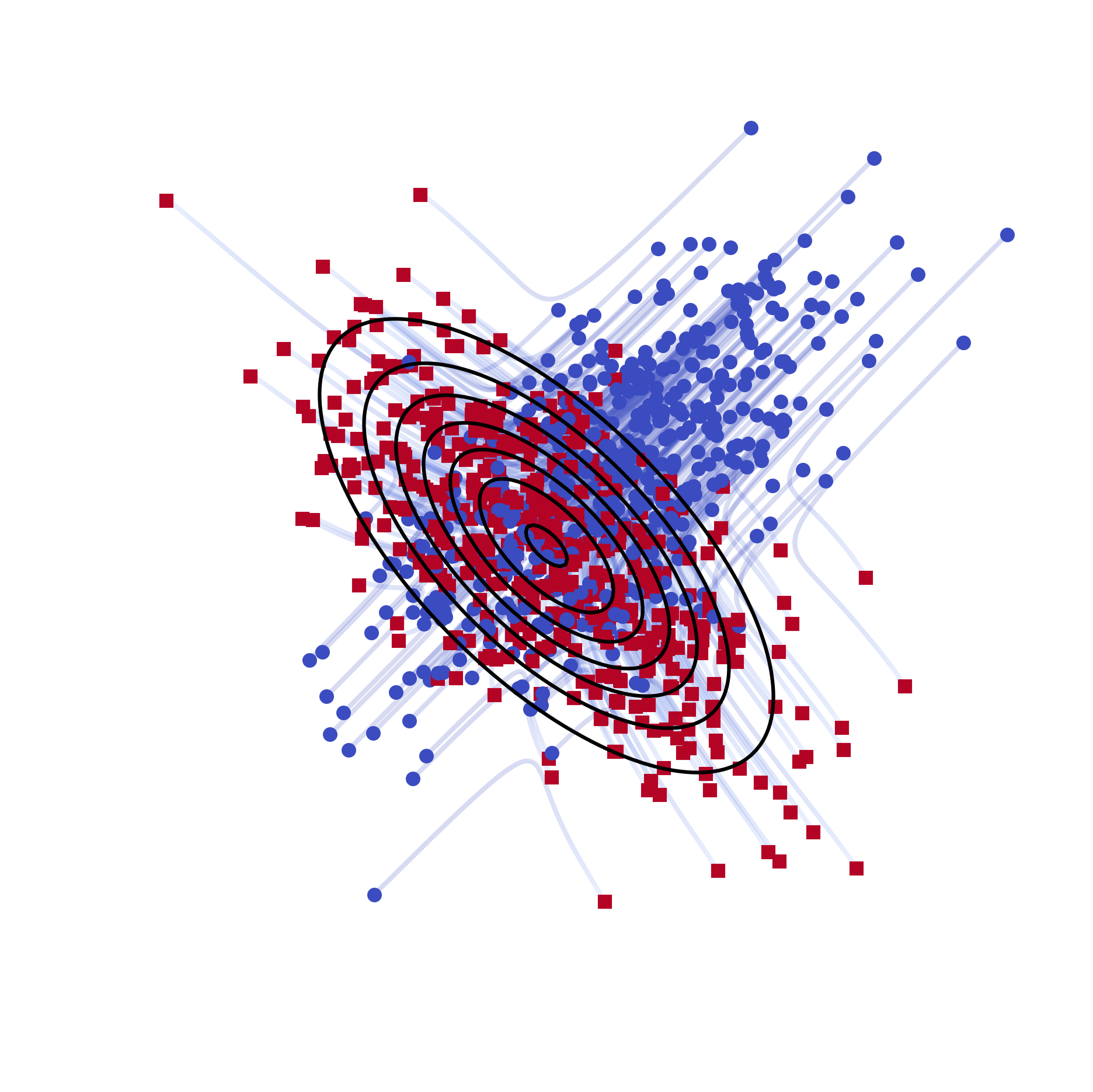}%
    \includegraphics[height=2.1cm]{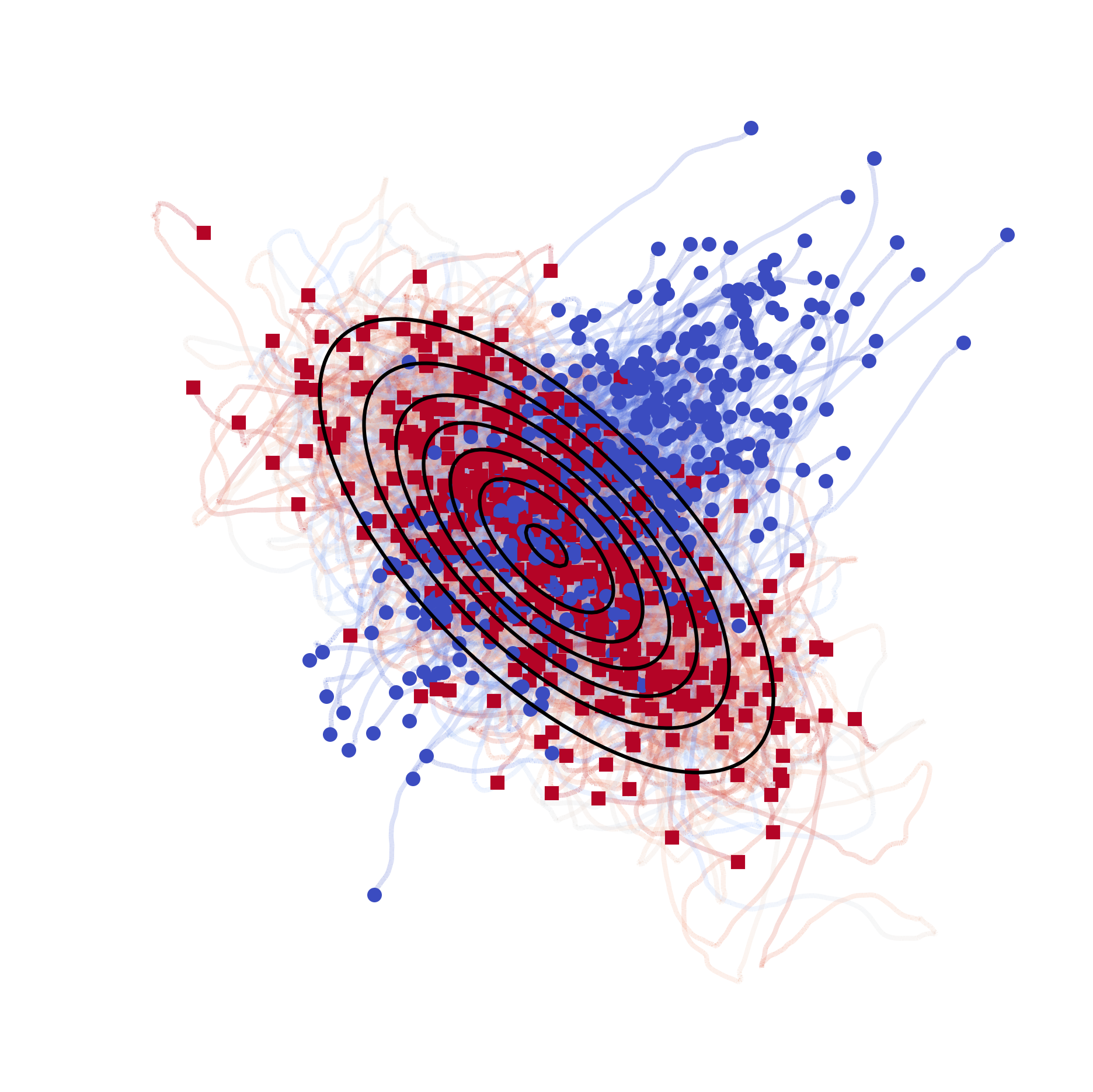}%
    \includegraphics[height=2.1cm]{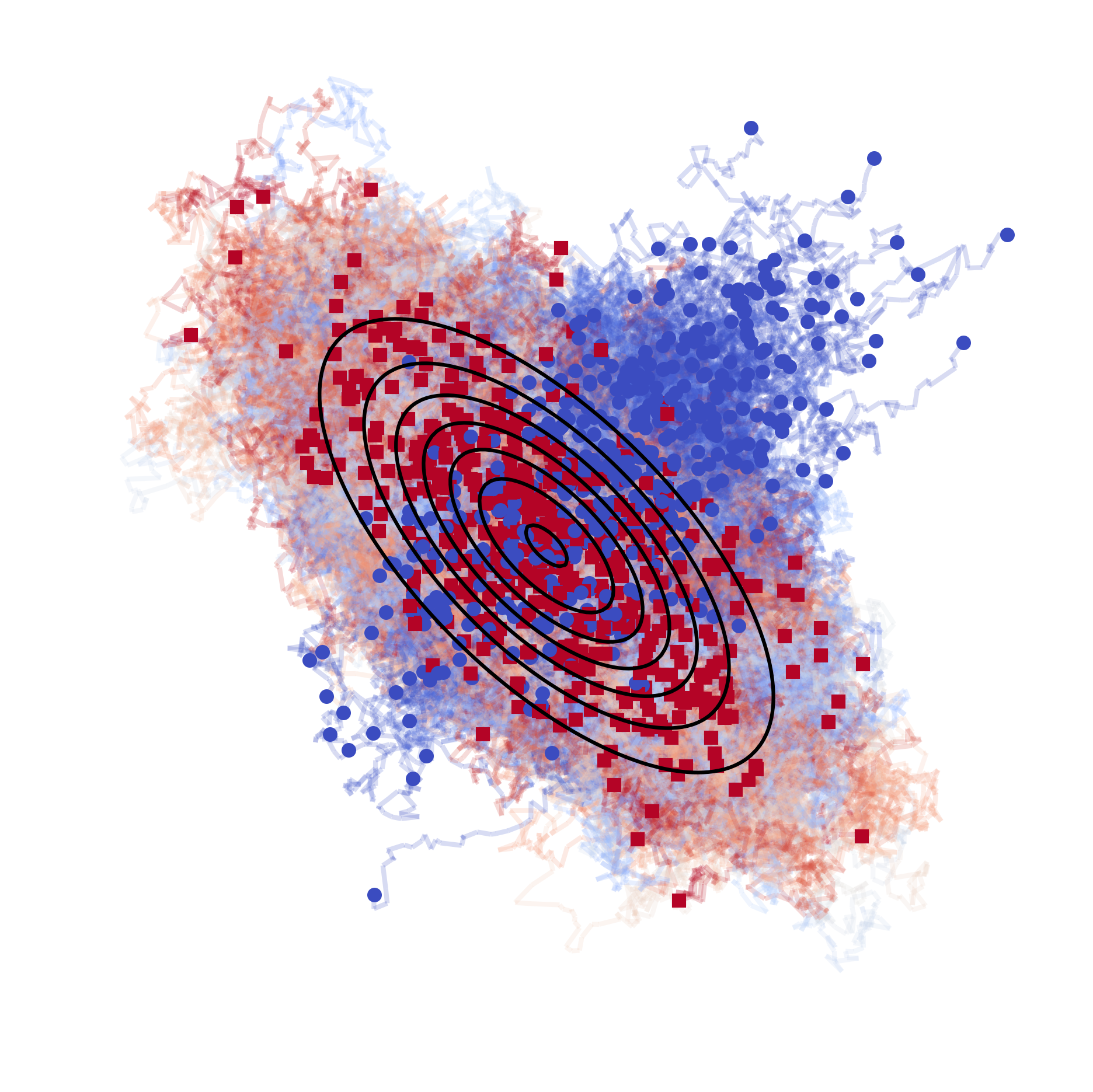}%
    \includegraphics[height=2.1cm]{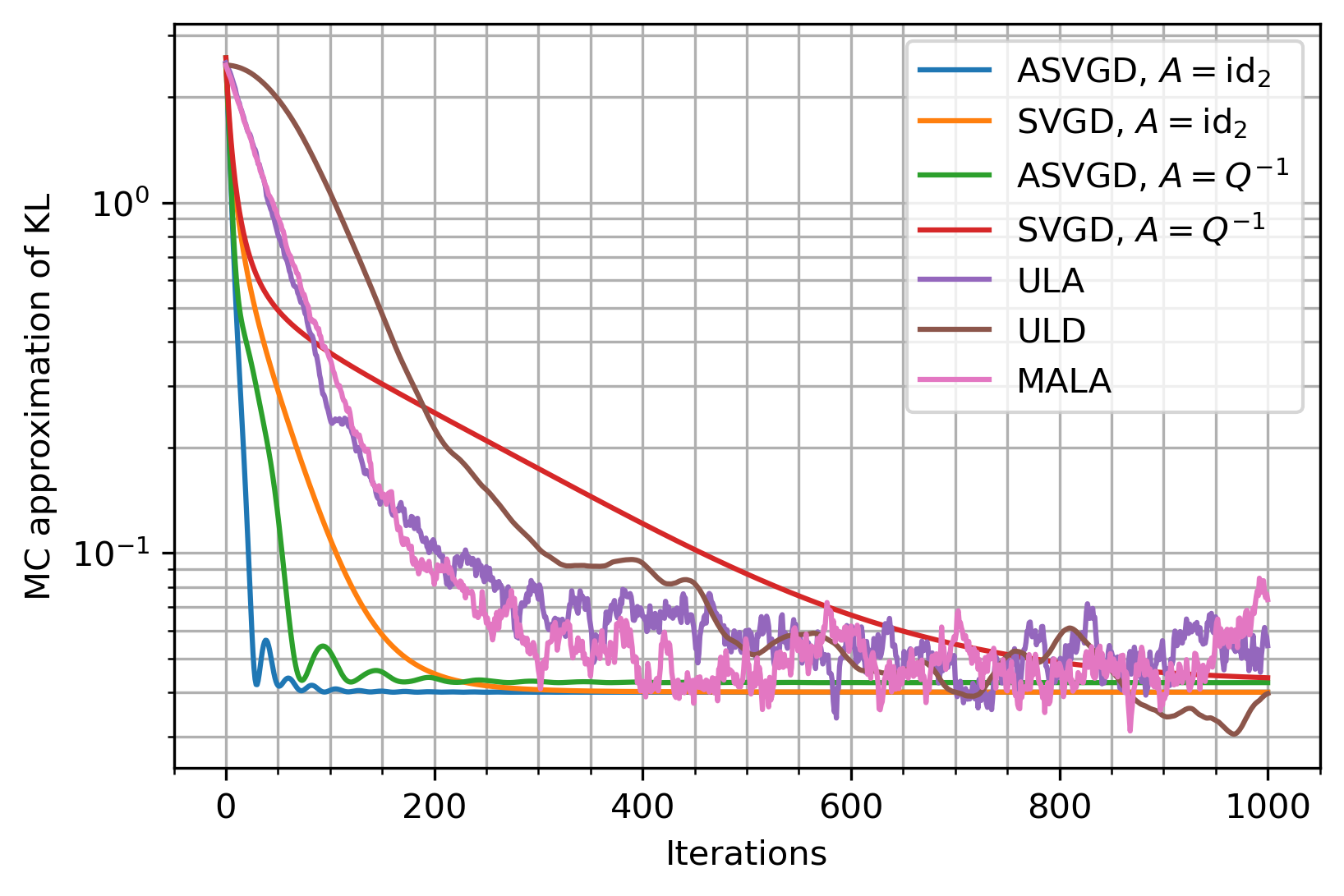}
    \caption{Particle trajectories of ASVGD, SVGD, with the generalized bilinear kernel, MALA, and ULD (from left to right) and the Monte-Carlo-estimated KL divergence for two different choice of $A$. The potential is $f(x) = \frac{1}{2} x^{\tT} Q x$, with $Q = [[3, -2], [-2, 3]]$ and the particles are initialized from a Gaussian distribution with mean $[1, 1]^{\tT}$ and covariance $[[3, 2], [2, 3]]$.}
\end{figure}\vspace*{-0.7cm}

\begin{figure}[t]
    \begin{subfigure}{\textwidth}
    \includegraphics[width=0.25\linewidth]{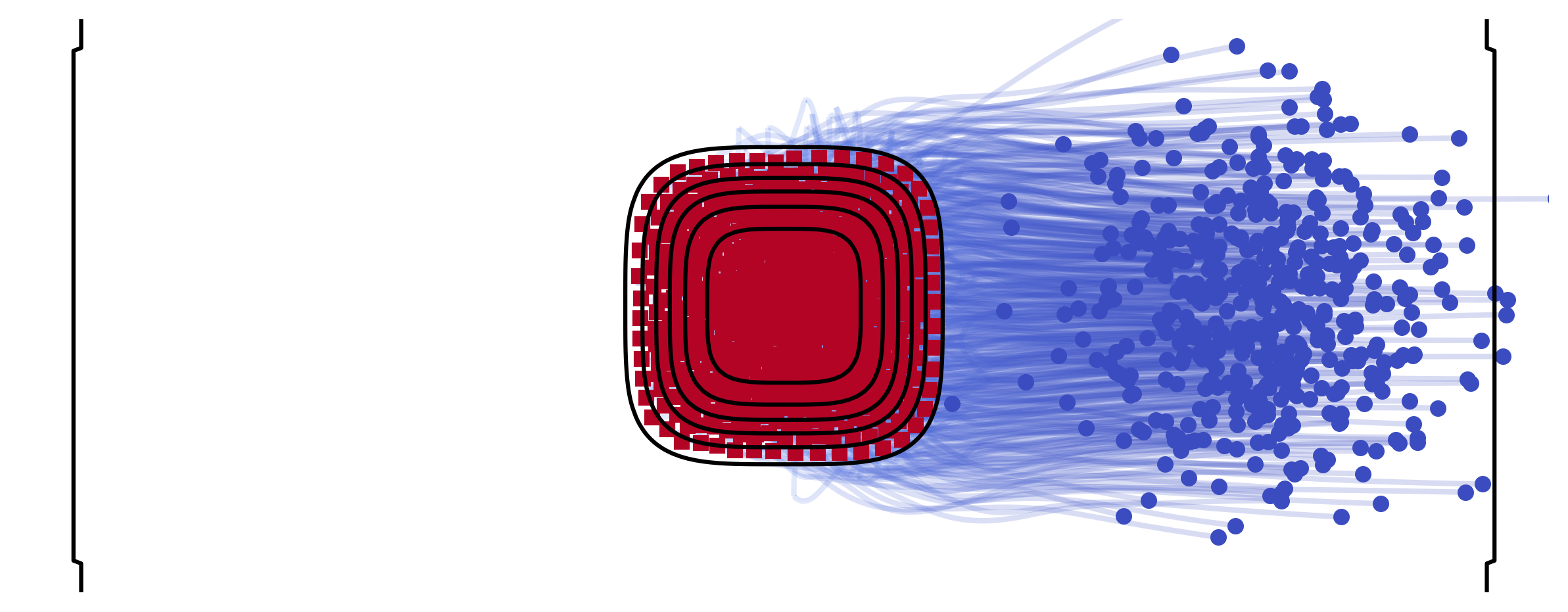}%
    \includegraphics[width=0.25\linewidth]{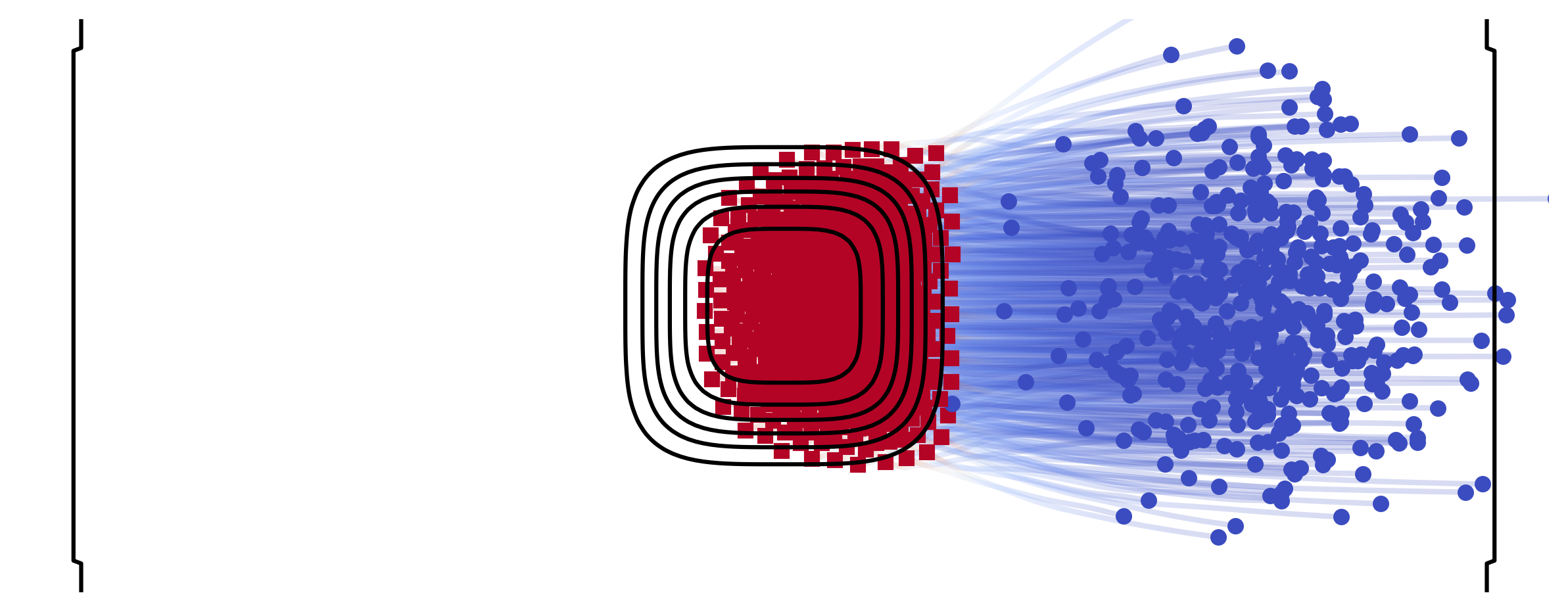}%
    \includegraphics[width=0.25\linewidth]{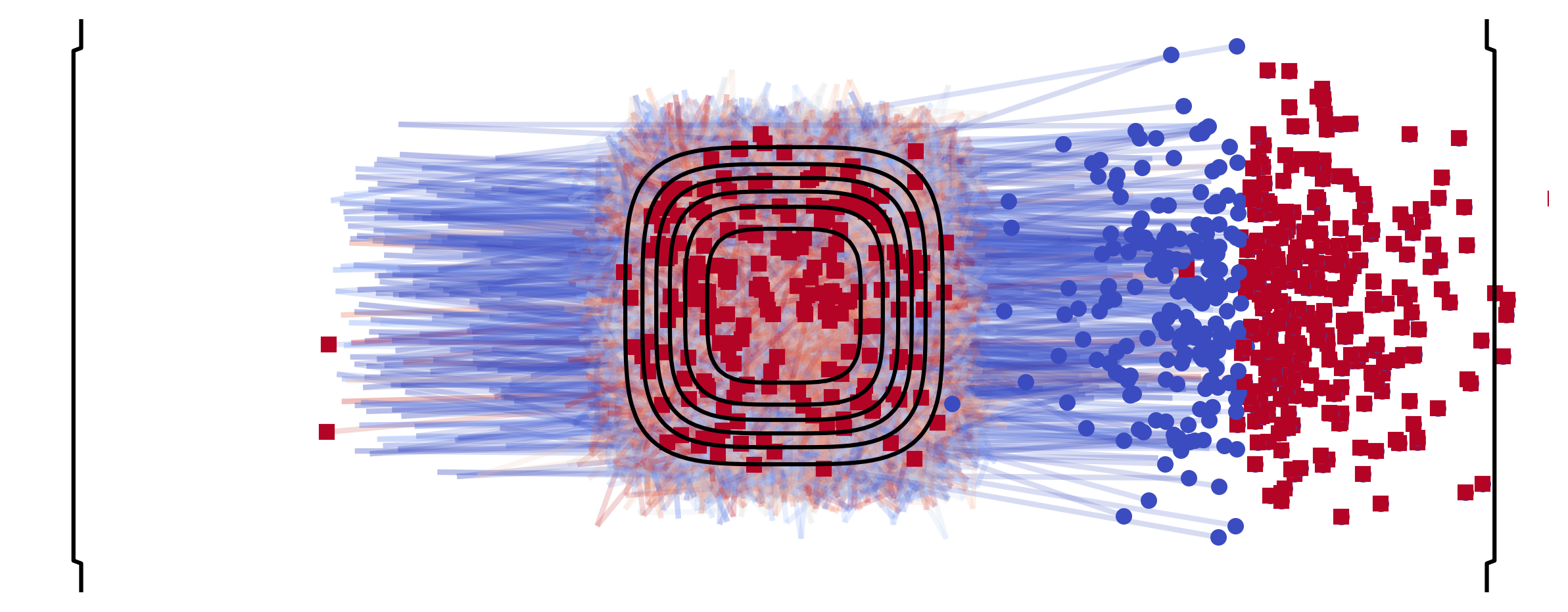}%
    \includegraphics[width=0.25\linewidth]{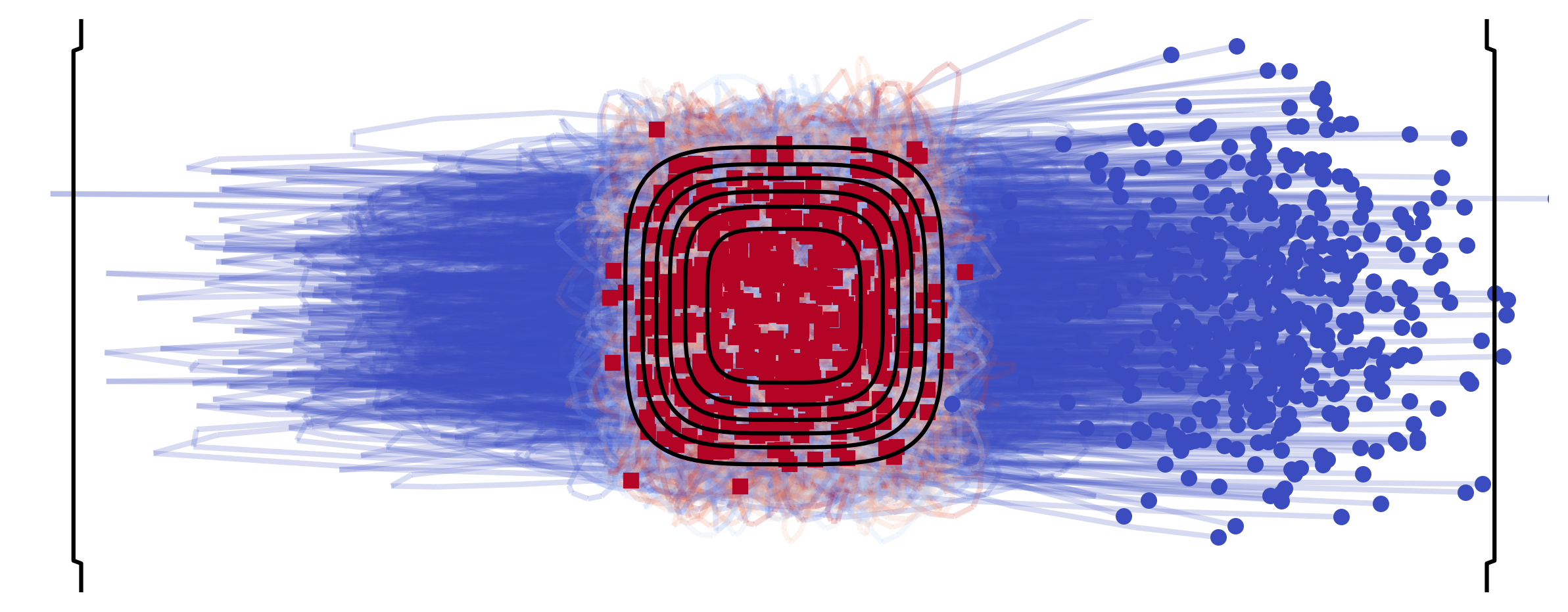}
    \includegraphics[width=0.25\linewidth]{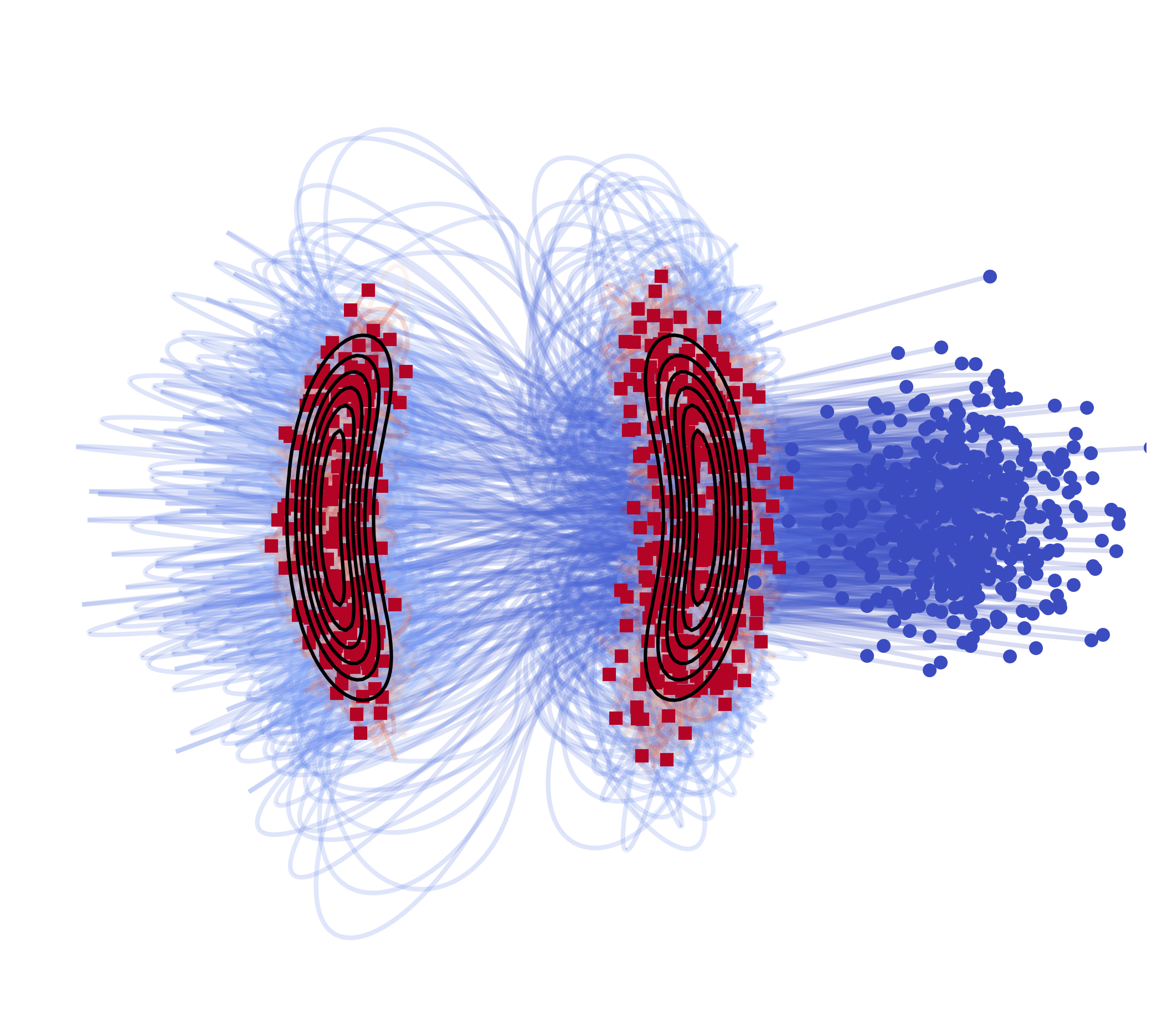}%
    \includegraphics[width=0.25\linewidth]{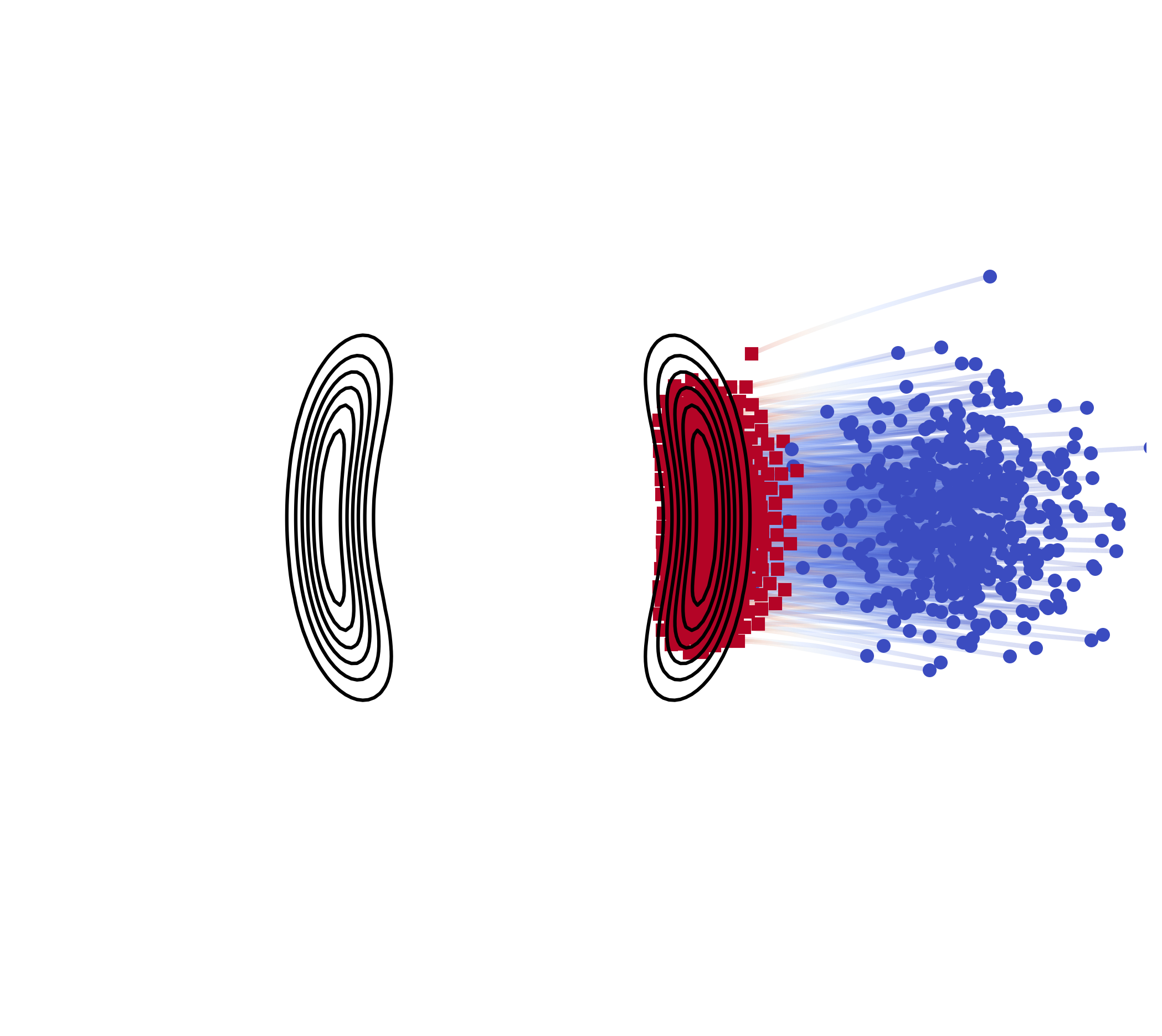}%
    \includegraphics[width=0.25\linewidth]{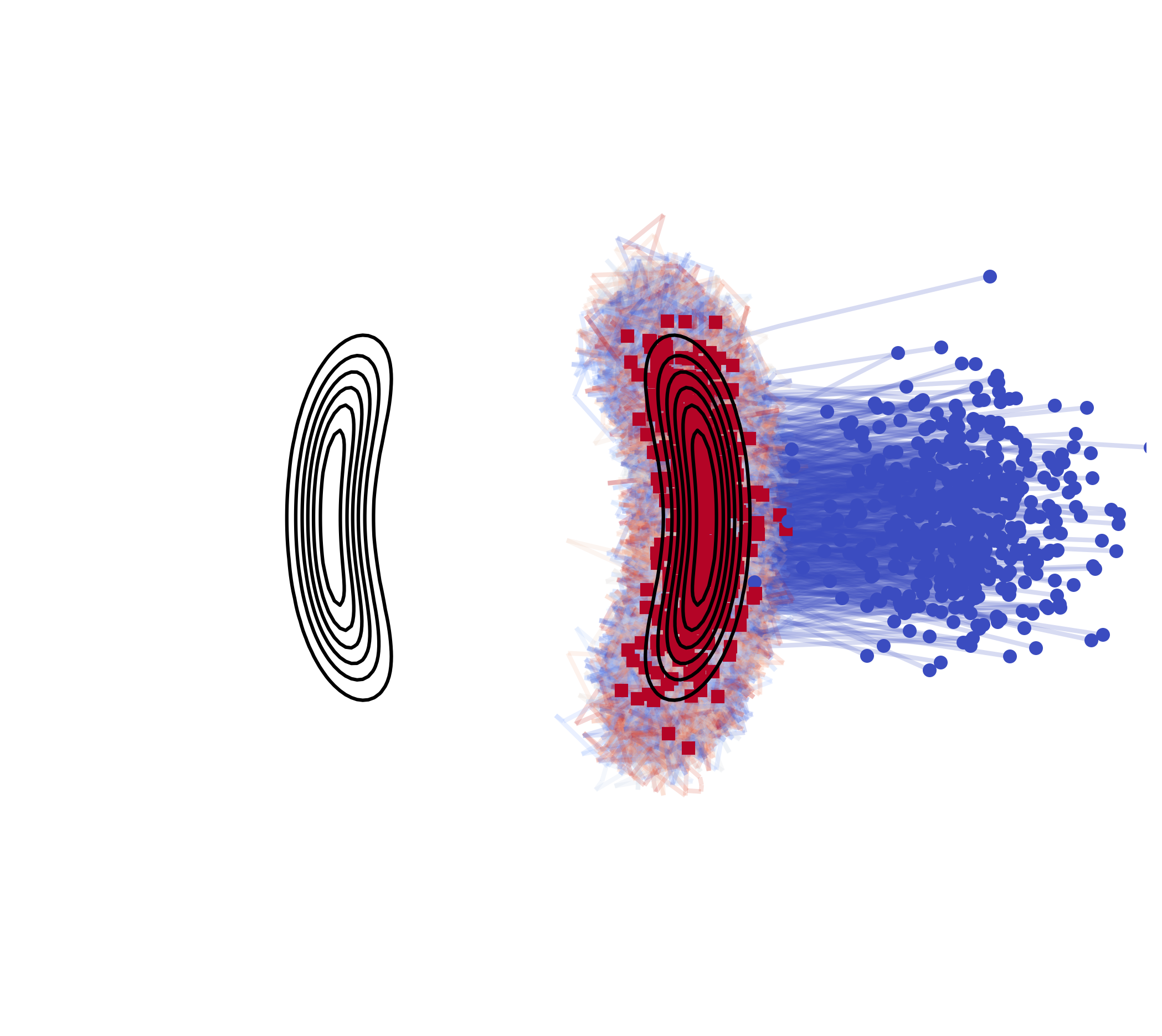}%
    \includegraphics[width=0.25\linewidth]{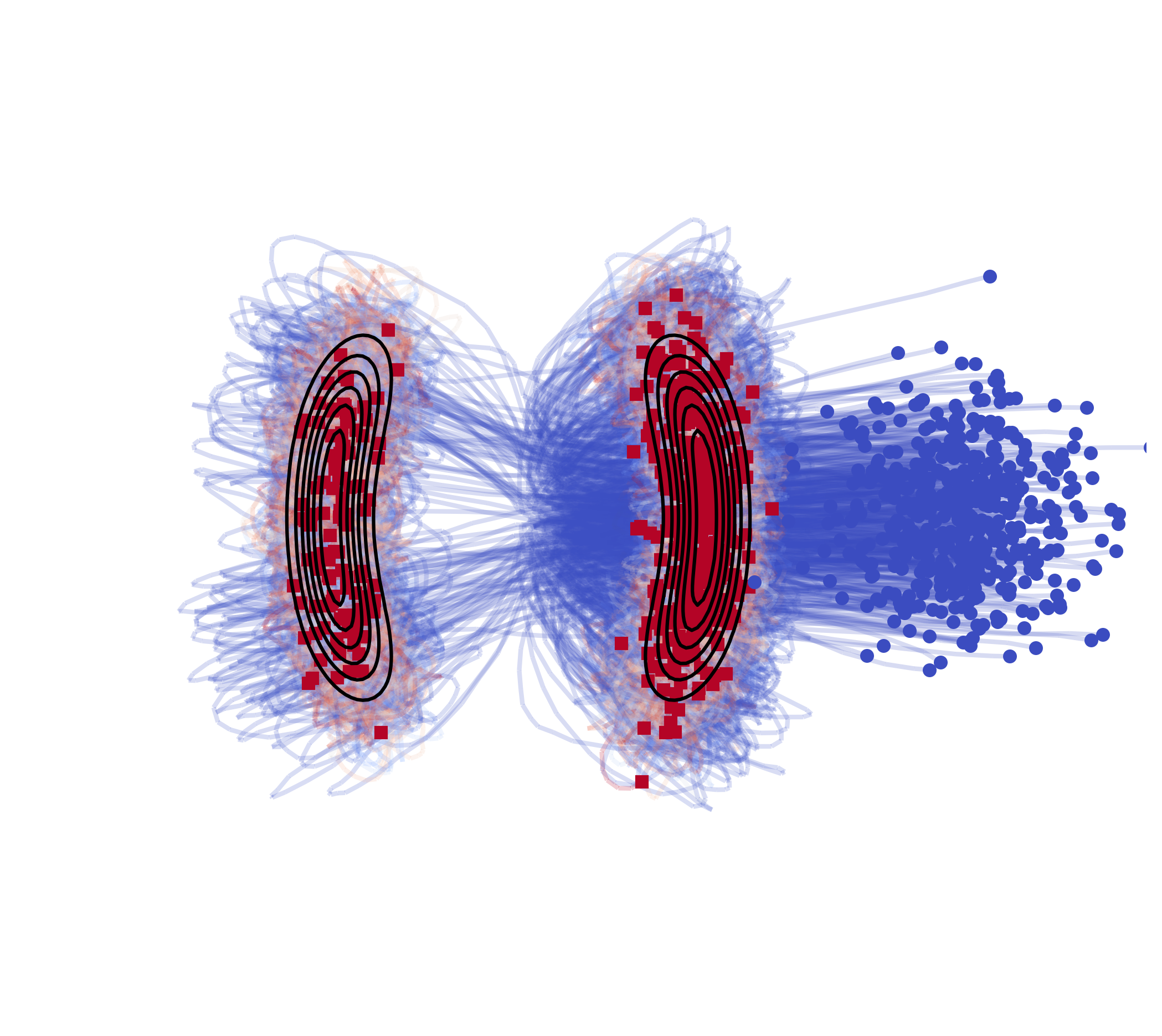}
    \includegraphics[width=0.25\linewidth]{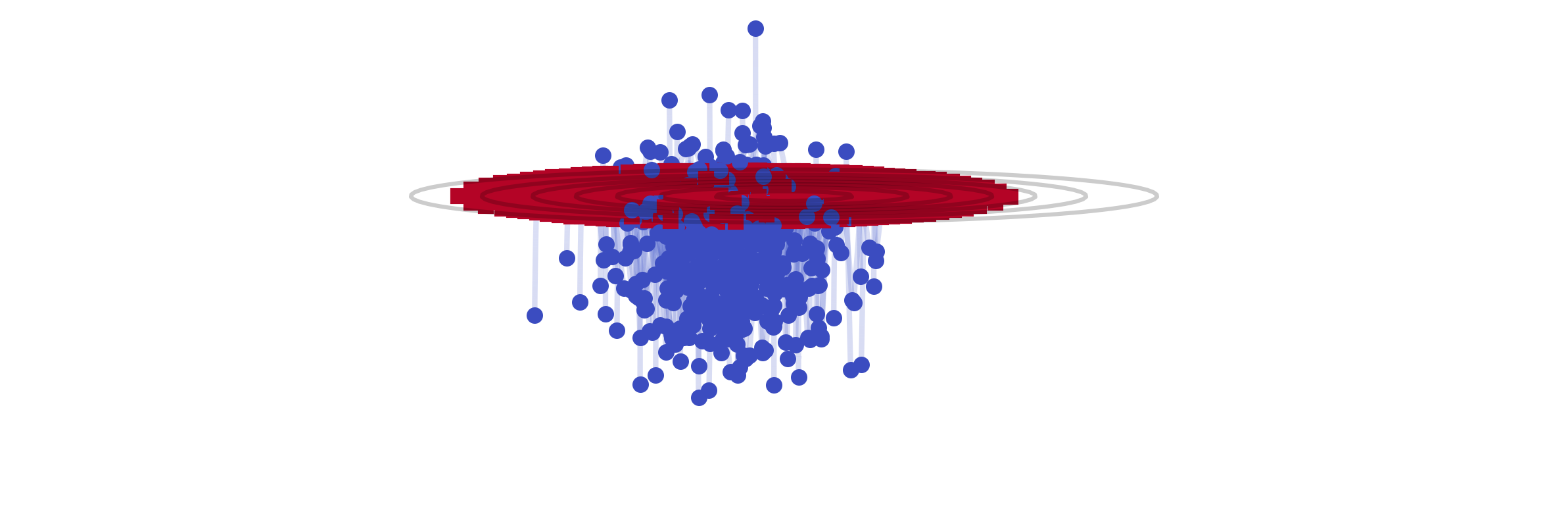}%
    \includegraphics[width=0.25\linewidth]{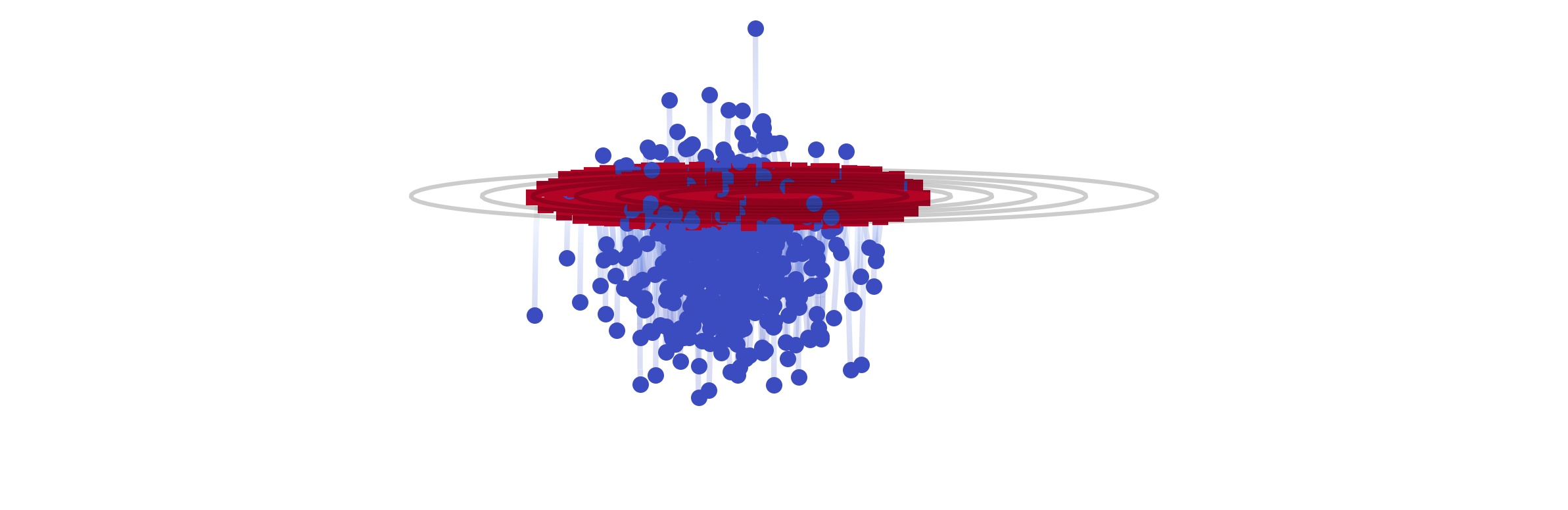}%
    \includegraphics[width=0.25\linewidth]{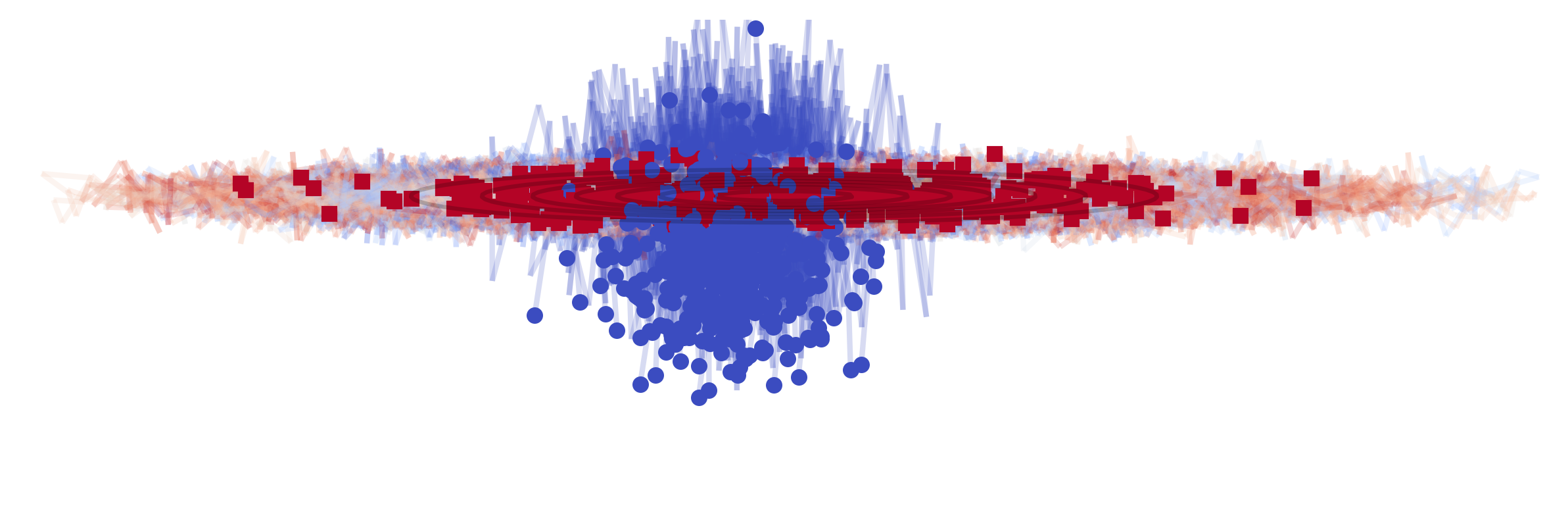}%
    \includegraphics[width=0.25\linewidth]{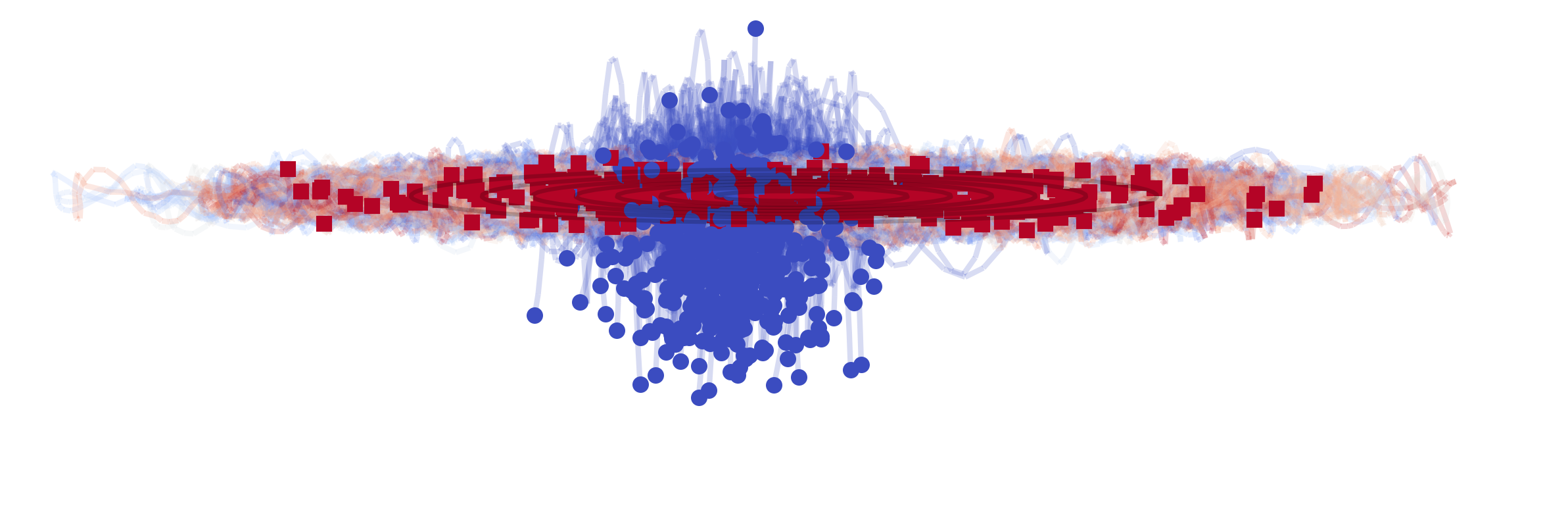}
    \subcaption{Particle trajectories of ASVGD, SVGD, with the Gaussian kernel, MALA, and ULD (from left to right) for the convex, non-Lipschitz potential $f(x, y) = \frac{1}{4}(x^4 + y^4)$ (top), the double bananas target from \cite{WL2022} (middle) and an anisotropic Gaussian target with mean $[1, 1]^{\tT}$ and covariance matrix $Q = \diag(10, 0.05)$ (bottom).}
    \end{subfigure}
    \begin{subfigure}{\textwidth}
    \includegraphics[width=0.33\linewidth]{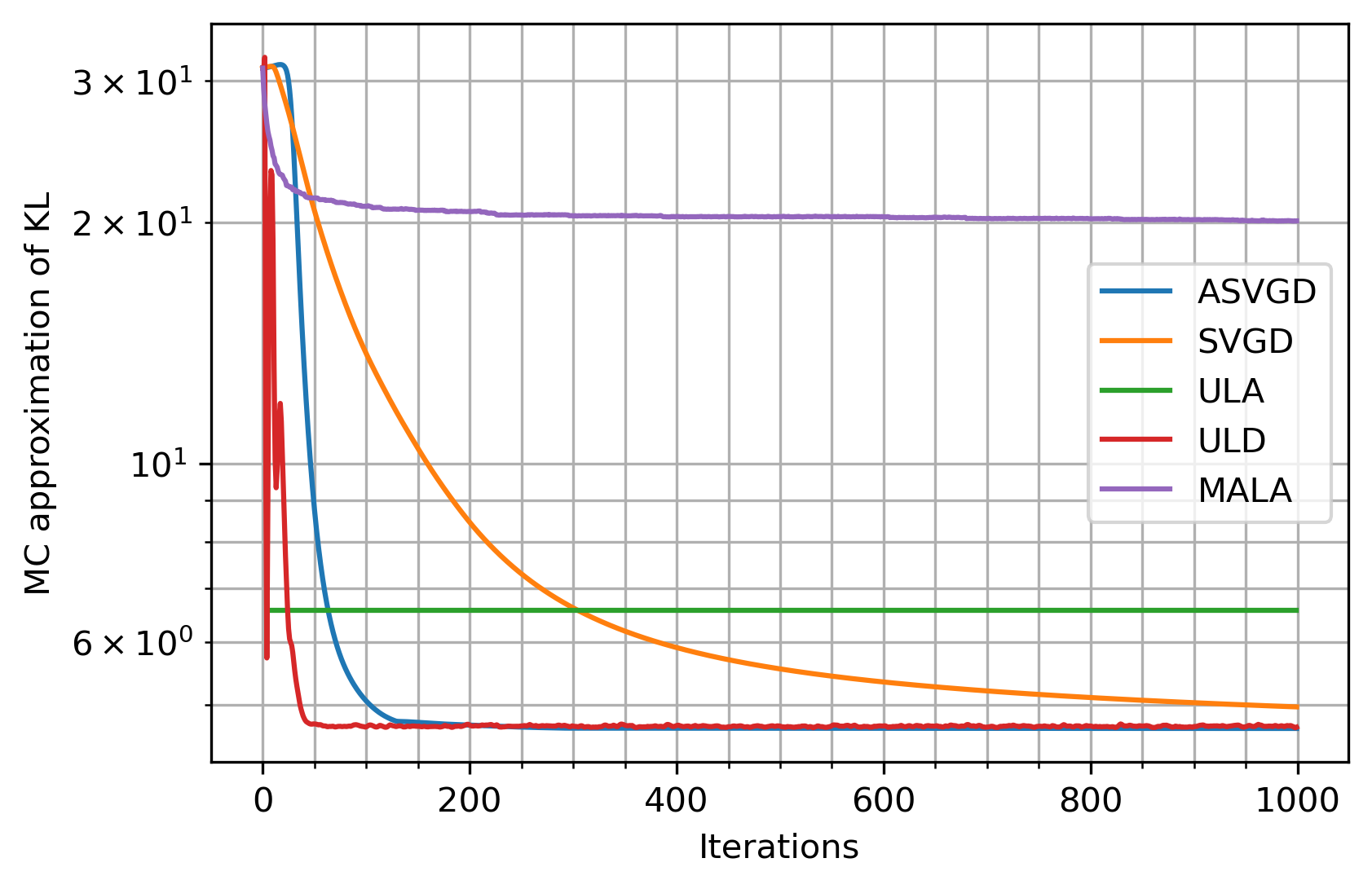}%
    \includegraphics[width=0.33\linewidth]{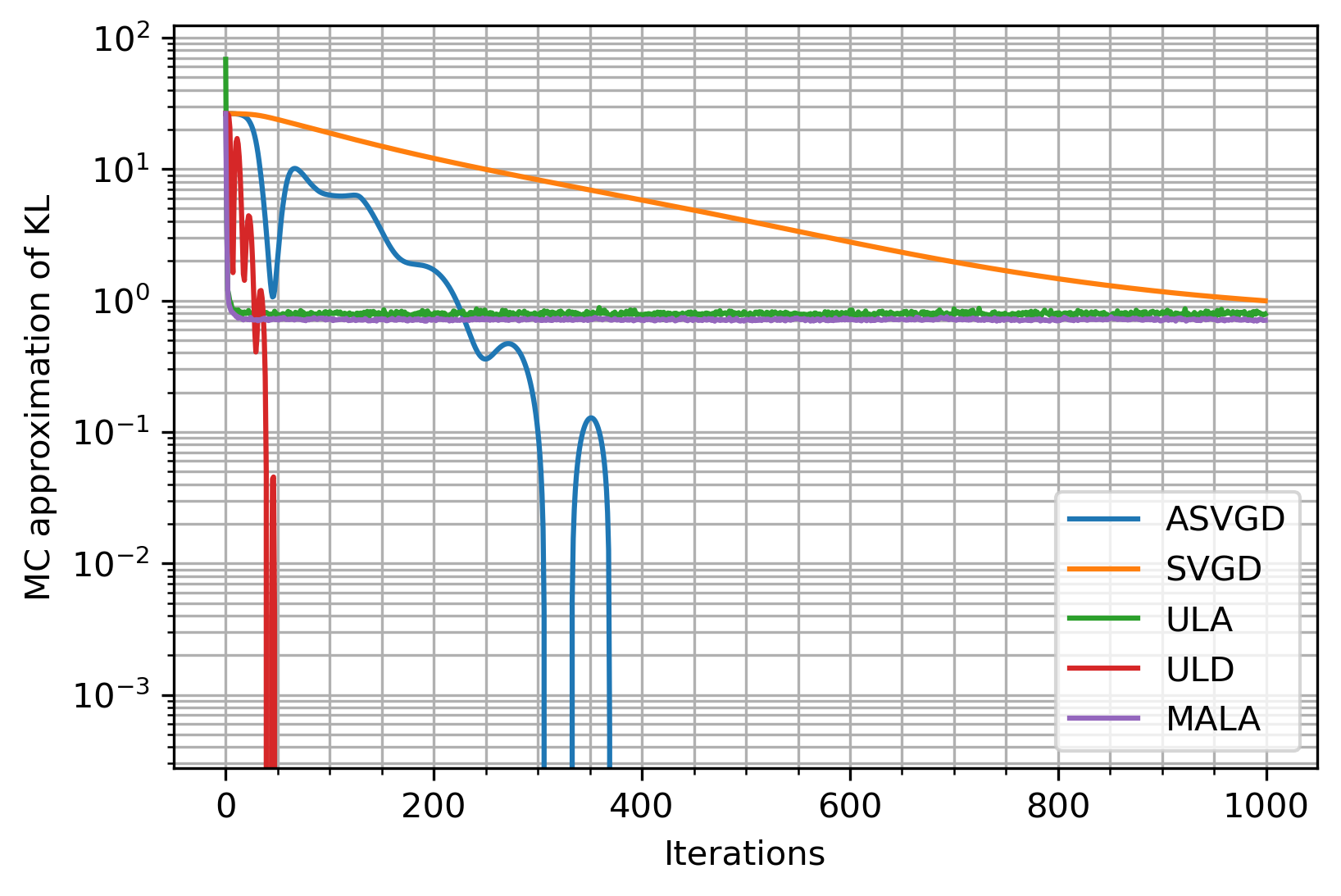}%
    \includegraphics[width=0.33\linewidth]{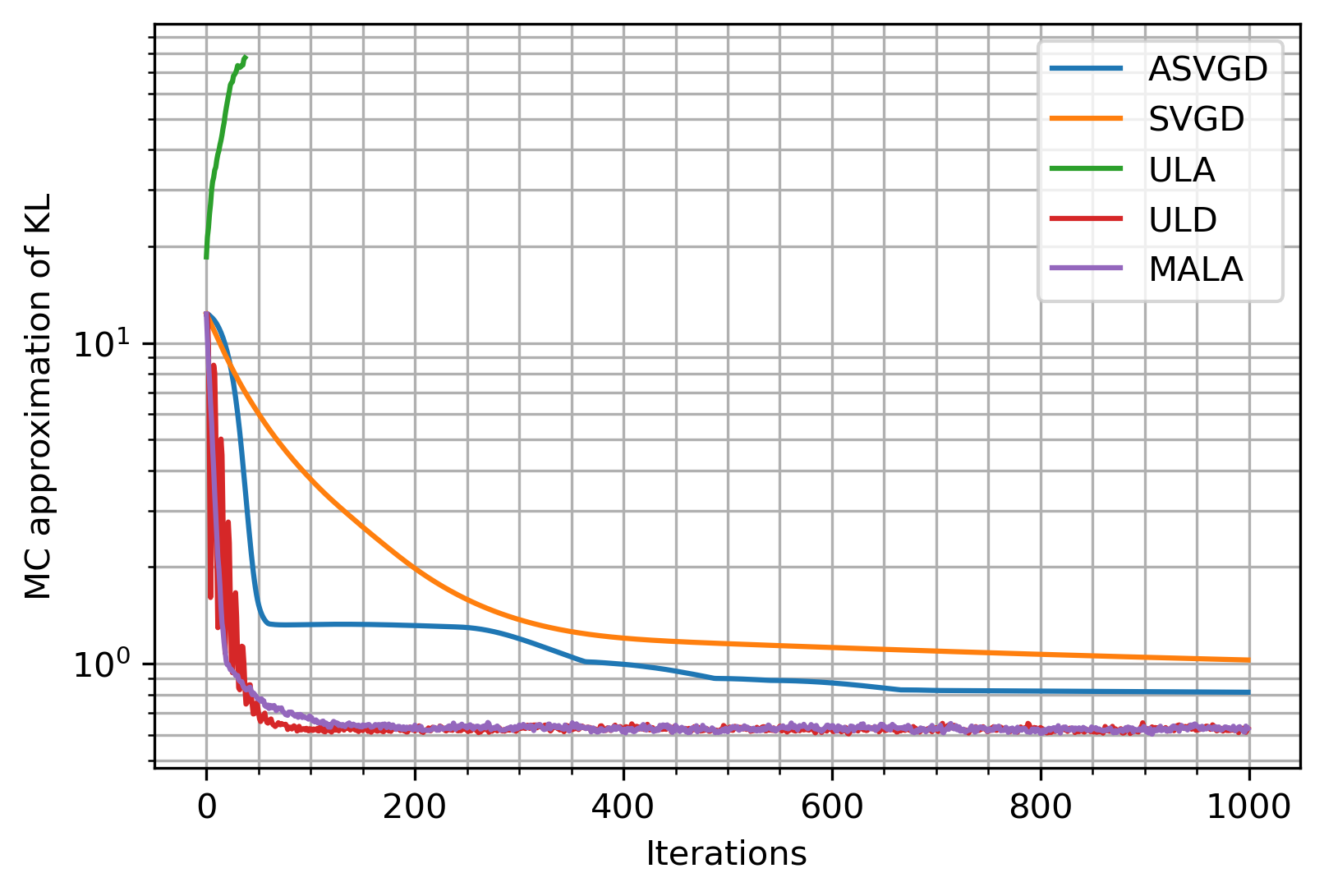}
    \subcaption{Monte-Carlo estimation of the KL divergence to the target for three targets above (left to right).}
    \end{subfigure}
    \caption{Comparing ASVGD to other sampling algorithms.
    For the double bananas target, we choose a constant high damping $\beta = 0.985$, for the other targets, we use the speed restart and the gradient restart.
    The initial particles are drawn from unit normal distributions with means $[0, 5]^{\tT}$, $[0, 7]^{\tT}$, and $[0, 0]^{\tT}$, respectively.}
    \label{fig:nonLip}
\end{figure}

\textit{Gaussian kernel.}
We observe that SVGD and ASVGD have lower variance.
Another advantage of our algorithm being deterministic is that particles arrange neatly along the level lines of the density, especially for small bandwidths, as opposed to the randomness of ULA, ULD, and MALA.
This is most pronounced in the first example in \figref{fig:nonLip}.
We also observe that the potential not being Lipschitz continuous leads to a high rejection rate, which slows down MALA considerably.
As observed for the anisotropic Gaussian target, ULA needs small step sizes to avoid degenerating, while ASVGD, as SVGD, works with large step sizes. Lastly, the momentum of ASVGD can improve exploration: only the second order in time methods, ASVGD and ULD, explore both modes and do not just concentrate on one banana of the target distribution. Note that the potential of the double bananas target is non-convex and non-smooth.

\section{Conclusion and future directions}
In this paper, we introduced ASVGD, an accelerated variant of SVGD, which can harness the advantages of momentum methods.
Our method remains score-estimation-free by using a momentum variable in the particle approximation. 
Further directions of interest include finding the optimal damping parameters $\alpha_k$ based on the convexity of the energy functional in metric spaces, and identifying the optimal kernel functions. We will also explore the connection between ASVGD and transformer architectures \cite{CACP2025}.

\begin{credits}
\subsubsection{\ackname}
W. Li’s work is supported by the AFOSR YIP award No.~FA9550-23-10087, NSF RTG: 2038080, and NSF DMS-2245097. 

\subsubsection{\discintname}
The authors have no competing interests to declare that are
relevant to the content of this article.
\end{credits}

\clearpage
%
%
\bibliographystyle{splncs04}
\bibliography{Bibliography}

\end{document}